\documentclass[letterpaper, 10 pt, conference]{ieeeconf/ieeeconf}  

\IEEEoverridecommandlockouts                              

\usepackage{graphics} 
\usepackage{epsfig} 
\usepackage{times} 
\usepackage{amsmath} 
\usepackage{amssymb}  
\usepackage{cite} 
\usepackage{algorithm}
\usepackage{algpseudocode}
\DeclareGraphicsExtensions{.pdf,.png,.jpg}
\usepackage{booktabs}
\usepackage{xcolor}
\usepackage{bm}
\usepackage{multirow}
\usepackage{siunitx} 

\algrenewcommand\algorithmicrequire{\textbf{Require:}}
\algrenewcommand\algorithmicensure{\textbf{Ensure:}}

\title{\LARGE \bf
    Risk-Aware Reinforcement Learning with Bandit-Based \\Adaptation for Quadrupedal Locomotion
}

\author{%
  \makebox[0.2\linewidth][l]{Yuanhong Zeng$^*$}%
  \makebox[0.2\linewidth][r]{Anushri Dixit$^\dagger$}
\thanks{$^*$Yuanhong Zeng is with the Department of Electrical and
Computer Engineering, University of California Los Angeles, Los Angeles,
CA, USA {\tt\small yuanhongzeng@ucla.edu}}%
\thanks{$^\dagger$Anushri Dixit is with the Department of Mechanical and Aerospace Engineering, University of California Los Angeles, Los Angeles,
CA, USA {\tt\small anushridixit@ucla.edu}}%
}

\begin{document}

\maketitle
\thispagestyle{empty}
\pagestyle{empty}

\begin{abstract}

In this work, we study risk-aware reinforcement learning for quadrupedal locomotion. Our approach trains a family of risk-conditioned policies using a Conditional Value-at-Risk (CVaR) constrained policy optimization technique that provides improved stability and sample efficiency. At deployment, we adaptively select the best performing policy from the family of policies using a multi-armed bandit framework that uses only observed episodic returns, without any privileged environment information, and adapts to unknown conditions on the fly. Hence, we train quadrupedal locomotion policies at various levels of robustness using CVaR and adaptively select the desired level of robustness online to ensure performance in unknown environments. We evaluate our method in simulation across eight unseen settings (by changing dynamics, contacts, sensing noise, and terrain) and on a Unitree Go2 robot in previously unseen terrains. Our risk-aware policy attains nearly twice the mean and tail performance in unseen environments compared to other baselines and our bandit-based adaptation selects the best-performing risk-aware policy in unknown terrain within two minutes of operation. 


\end{abstract}

\section{Introduction}

Quadrupedal robots can traverse a variety of unstructured terrains, enabling tasks such as mapping hard-to-reach areas, detecting life signs, and reducing human exposure to hazardous environments. Early locomotion controllers utilized model predictive control~\cite{di2018dynamic} which requires accurate dynamics models that are often hard to obtain. With modern simulators such as Isaac Sim~\cite{mittal2023orbit} that can simulate hard-to-model dynamics under different conditions, quadrupeds can learn to walk within minutes using model-free reinforcement learning (RL) and massive parallelization~\cite{rudin2022learning}.

A significant challenge in deploying policies trained in simulation onto real hardware is that the shift between training and deployment conditions may cause a difference in the performance expected versus what is attained in the real world. A common mitigation strategy is domain randomization~\cite{tobin2017domain}, but selecting parameters and ranges that both stabilize training and produce good deployment performance requires substantial manual effort. Risk-aware RL methods offer another way to mitigate 
sim-to-real shift by emphasizing performance in the tail of the training distribution rather than maximizing expected returns~\cite{shi2024robust}. However, these methods can be overly conservative and difficult to train, as they require specifying a desired risk (or robustness) level a priori. In practice, because the deployment environment is unknown, any chosen risk level is likely to be miscalibrated, leading to behaviors that are either too optimistic or too conservative.

We address these issues using CVaR-constrained optimization: rather than maximizing CVaR directly, we maximize expected return subject to a constraint on the CVaR of the return. This formulation allows us to use standard estimators such as Generalized Advantage Estimation~\cite{schulman2015high} and Proximal Policy Optimization (PPO)~\cite{schulman2017proximal}. We enforce the CVaR constraint using a Lagrangian formulation, allowing the Lagrange multiplier adaptively balance constraint satisfaction and return maximization, which improves training stability. Prior work on CVaR-constrained RL~\cite{ying2022towards} has shown that satisfying the constraint yields safety guarantees in perturbed environments.

\begin{figure}[t]
\centering
\includegraphics[width=\columnwidth]{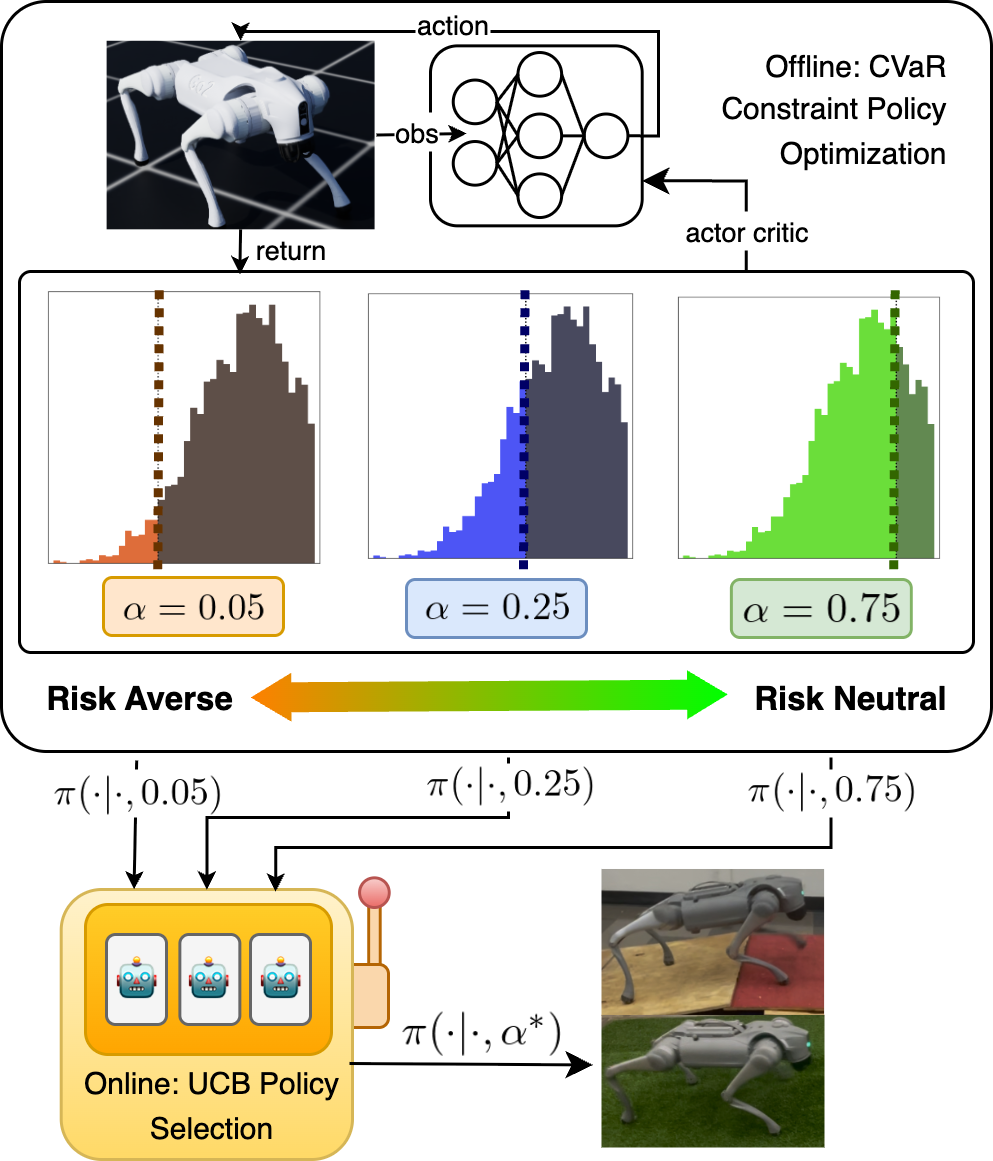}
\caption{Overview: We train multiple policies using CVaR-constrained policy optimization, where each critic focuses on a different tail of the return distribution, resulting in policies with varying levels of risk awareness. An upper confidence bound (UCB) bandit is then used to adaptively select the appropriate risk level online. The bandit selects the best performing policy over time as the robot interacts with the environment repeatedly.}
\label{fig:anchor}
\end{figure}

As shown in Figure~\ref{fig:anchor}, we train a family of CVaR-constrained reinforcement learning policies at different risk-levels, $\alpha \in (0,1)$. A risk-neutral policy may perform well in simple, flat ground but may fail on undulating terrain. On the other hand, a policy that is risk-averse may be successful over rocky terrain but will be too slow or conservative on flat ground. Because deployment conditions can range from near-nominal to unstructured, we introduce an online policy selection mechanism using a multi-armed bandit that adaptively chooses the most appropriate risk-level for the current environment.

We evaluate our approach in both simulation and hardware. In simulation, we test across eight environments with variations in parameters, mechanical properties, unexpected contacts, and sensor noise. On a Unitree Go2, we validate that risk-aware behavior improves task success. Our method achieves nearly twice the mean and tail performance of a PPO baseline trained in the same environment when deployed under disturbances, and the bandit selector converges to the best policy for an unknown test environment within two minutes of wall-clock time. Our contributions can be summarized as follows:
\begin{itemize}
\item We propose a framework for learning risk-aware policies via CVaR-constrained RL.
Our RL algorithm augments PPO with a clipped-Lagrangian to account for the CVaR constraint, improving training stability and sample efficiency.
\item We present an online Upper Confidence Bound (UCB)-based multi-armed bandit that selects the risk level online to match and adapt to unknown environment conditions.
\item We conduct comprehensive evaluation in simulation and on a Unitree Go2 Edu robot, demonstrating robust behavior under diverse disturbances and rapid convergence of the selection mechanism to the best policy.
\end{itemize}

\section{Related Works}

\subsection{Distributional RL}

Most risk-aware RL methods optimize a tail-risk objective directly. For example,~\cite{shi2024robust} trains a quadruped by maximizing the CVaR of return, and~\cite{schneider2024learning} spans a spectrum of risk attitudes by optimizing various distortion risk measures, including CVaR and the Wang metric~\cite{wang2000class}. These approaches typically rely on distributional RL~\cite{bellemare2017distributional, dabney2018distributional, dabney2018implicit} to model the return distribution and design critics whose advantages reflect the chosen risk functional. More recent work uses return capping to avoid explicit distribution modeling, yielding faster and more stable training~\cite{mead2025return}. A drawback of tail-objective optimization is that the critic downweights high-return trajectories, which can reduce sample efficiency and increase the required environment interaction.

\subsection{Risk-Aware, Constrained RL}

We build our work based on the constrained RL~\cite{schulman2015trust, achiam2017constrained, miryoosefi2019reinforcement} framework. We highlight two benefits of the constrained approach. First, constrained RL formulation gives better performance guarantee. Ying et al.~\cite{ying2022towards} show that the performance gap when deployed in an environment with different transition probability can be can be decreased by increasing the value at the worst state, and CVaR of return can be seen as a surrogate, which is bounded below when the constraint is satisfied. Second, the gradient signal comes from both the expected return object and the CVaR constraint penalty. This yields better sample efficiency and training results compared to previously discussed methods only optimizes tailed return. 


\subsection{Online Adaptation and Learning}

Most state-of-the-art locomotion methods adaptively adjust policy behavior. A common approach is to condition the policy on environment features estimated from different modalities such as depth vision~\cite{yang2023neural}, proprioception~\cite{kumar2021rma}, or LiDAR~\cite{yang2021learning}. However, these methods usually assume access to privileged information during training (e.g., accurate height maps) and still depend on carefully designed training environments. An alternative line of research \cite{shi2024robust, schneider2024learning}, conditions policy behavior on the return distribution. Our work is closely related: we employ a UCB-based bandit to explore the return distribution and select actions that maximize its upper confidence bound. Bandit algorithms \cite{tamkin2019distributionally, mnih2008empirical, contextual}, provide a principled way to balance exploration and exploitation with theoretical regret guarantees \cite{bandit2019}, adapting directly from observed returns instead of relying on hand-crafted features. Compared to heuristic or feature-matching methods, they are more robust to environment shifts, require less privileged information. 

\section{Preliminaries: Conditional Value-at-risk}

For stochastic optimization problems, constraints can be reformulated by a commonly used risk measure called the \textit{Value-at-Risk} (VaR). Given a confidence level $\alpha \in (0,1)$, $\mathrm{VaR}_{\alpha}$ denotes the ${\alpha}$-quantile value of the return variable $R$ and is defined as, 
\begin{align*}
    \text{VaR}_{\alpha}(R) := \inf \{ \eta\,|\, \mathbb{P}(R \leq \eta) \geq \alpha\}.
\end{align*}
It follows that $
    \text{VaR}_{\alpha}(R)\geq 0 \implies \mathbb{P}(R \leq 0)\leq \alpha.$
However, VaR is generally nonconvex and hard to compute. We now introduce a convex and monotonic risk measure called the conditional value-at-risk.

The {\em conditional value-at-risk}, $\mathrm{CVaR}_{\alpha}$, measures the expected loss in the ${\alpha}$-tail below the threshold $\mathrm{VaR}_{\alpha}$. $\mathrm{CVaR}_{\alpha}$ is computed as{~\cite{rockafellar2000optimization},
\begin{equation}
\label{eq:cvar}
\begin{aligned}
    \mathrm{CVaR}_{\alpha}(R):=&\sup_{\eta \in \mathbb{R}}\mathbb{E}\Bigg[\eta - \frac{(\eta-R)^{+}}{\alpha}\Bigg],
\end{aligned}
\end{equation} 
}
where $(\cdot)^{+}:=\max\{\cdot, 0\}$. A value of $\alpha \simeq 1$ corresponds to a risk-neutral case where $\mathrm{CVaR}_{\alpha\simeq 1}(R)\rightarrow\mathbb{E}[R]$ and a value of $\alpha \to 0^+$ is risk-averse.  
CVaR provides a convex lower bound of VaR, i.e.,
\begin{align}
    0\leq\small{\text{CVaR}_{\alpha}(R) \leq \text{VaR}_{\alpha}(R) \implies \mathbb{P}(R\leq 0)\leq \alpha.}
\end{align}
\section{Problem Statement}
\label{sec:problem}

We model quadruped velocity-tracking as a discounted Markov Decision Process (MDP) $M := (\mathcal{S}, \mathcal{A}, P, R, \gamma)$ with discount factor $\gamma\in(0,1)$. The state space $\mathcal{S}$ comprises base linear and angular velocities, normalized projected gravity vector, velocity command, joint positions, joint velocities, and past action vector. $\mathcal A$ is the action space which consists of target joint positions. $P$ is the state transition kernel and $R: \mathcal S\times \mathcal A \rightarrow \mathbb R$ is the reward function. We assume the rewards are bounded. For a stationary policy $\pi$ under MDP, $M$, the $\alpha$-tailed return, $\alpha \in (0,1]$, is defined as,
\begin{equation*}
    J(\pi;\alpha, M):= \text{CVaR}_{\alpha}\!\left[\sum_{t=0}^{\infty}\gamma^t R(s_t,a_t)\right]
\end{equation*}
where $a_t\sim\pi(\cdot\mid s_t)$ and $s_{t+1}\sim P(\cdot|s_t,a_t)$.

We define $J(\pi) := J(\pi;1,M)$ to be the expected value of the sum of the discounted reward of policy $\pi$ when deployed in $M$, and $D(\xi) = \sum^\infty_{t=0} \gamma^t R(s_t,a_t)$ to be the sum of discounted reward of the trajectory $\xi = (s_0, a_1, s_1,\ldots)$. 

For a policy is deployed in an unknown environment, we define a perturbed MDP $\widehat M = (\mathcal S, \mathcal A, \widehat P, R, \gamma)$ with the same state space, action space, and reward but a shifted transition kernel $\widehat P$.

\textbf{Problem 1} For the MDP $\widehat M$, we seek a risk-conditioned policy $\pi_\theta(a|s, \alpha)$,  with risk parameter $\alpha \in (0, 1]$ and policy parameter $\theta$ by maximizes the $\alpha$-tailed return under an unknown transition kernel $\widehat P$.
\begin{equation}
\label{eq:problem1}
    \theta^\ast = \arg \max_\theta J(\pi_\theta; \alpha, \widehat M)
\end{equation}

\textbf{Problem 2} Suppose we have a finite set of policies $\{\pi_{\theta_k}(\cdot|\cdot, \alpha_k)\}^K_{k=1}$, each trained with a distinct risk level $\alpha_k$. We formulate online policy selection problem as a stochastic multi-arm bandit over episode problem~\cite{bandit2019}. At episode $e$, the learner selects an arm in $k \in \{1, \ldots, K\}$, executes $\pi_{\theta_k}$ for $T$ steps in $\widehat M$, and then observes the undiscounted episodic return $X_{k,e} =\sum^{T-1}_{t=0} R(s_t, a_t)$. For each arm $k$, we assume that $\{X_{k,e}\}_{e=1}^H$ are $H$ i.i.d draws from a stationary distribution with bounded mean and variance. The goal is to minimize regret defined as
\begin{equation}
\label{eq:problem2}
    \mathcal{R}(H;\alpha) := H\cdot\mathbb E X_{k^\ast} - \sum^H_{e=1} X_{k_e,e}
\end{equation}

where, $k_e$ is the arm selected at episode $e$, and $k^\ast = \arg\max_k \mathbb E X_{k}$.

\section{Method}




\subsection{CVaR Constrained Proximal Policy Optimization}

Unlike recent methods based on return capping\cite{mead2025return} and distributional RL~\cite{bellemare2017distributional, dabney2018implicit, ma2025dsac, shi2024robust} that attempt to solve Problem 1 by directly maximizing the CVaR of return, we propose to solve the CVaR constrained problem,

\begin{equation}
\label{eq:cvarconstraint}
    \begin{aligned}
    \max_\theta\quad&\mathbb E_{a_t\sim\pi(\cdot\mid s_t),s_{t+1}\sim \hat P(\cdot|s_t,a_t)} \left[\sum^\infty_{t=0}\gamma^t R(s_t,a_t)\right]\\
    \text{s.t}\quad& J(\pi;\alpha, \widehat M) \ge \beta.
    \end{aligned}
\end{equation}

Since samples from $\widehat P$ are not available, we use samples from $P$ to approximate the corresponding expectations. We first convert the constrained problem in~\eqref{eq:cvarconstraint} into an unconstrained form via a Lagrangian relaxation~\cite{bertsekas1997nonlinear} and using the CVaR definition in {\eqref{eq:cvar}},

\begin{equation}
\label{eq:cvarunconstraint}
\begin{aligned}
\max_{\lambda\ge 0}\;\min_{\theta,\eta}\; &L(\theta,\eta,\lambda)
:= -J(\pi_\theta) \\
&\quad + \lambda\!\left(\beta -\eta + \frac{1}{\alpha}\,\mathbb{E}_{\xi\sim \pi_\theta}\!\left[(\eta - D(\xi))^+\right] \right).
\end{aligned}
\end{equation}

\noindent To solve~\eqref{eq:cvarunconstraint}, we use the following gradients from~\cite{JMLR:v18:15-636},

\begin{align}
\label{eq:eta}
\nabla_{\eta} L &= -\lambda \;+\; \frac{\lambda}{\alpha}\, \mathbb{E}_{\xi \sim \pi_\theta}\mathbf{1}\{D(\xi) \le \eta\},\\
\label{eq:policy_update}
\nabla_{\theta} L &= -\,\mathbb{E}_{\xi\sim \pi_\theta}\!\big[\nabla_\theta\log P_\theta(\xi)\big] \bigg(D(\xi) - \frac{\lambda}{\alpha}\,(\eta - D(\xi))^+\bigg),\\
\label{eq:lambda}
\nabla_\lambda L &= \beta -\eta + \frac{1}{\alpha}\,\mathbb{E}_{\xi\sim \pi_\theta}\!\left[(\eta - D(\xi))^+\right].
\end{align}

\noindent Although Chow et~al.\ show that gradient descent using \eqref{eq:eta}-\eqref{eq:lambda} converges to locally optimal policies in~\cite{JMLR:v18:15-636}, in practice directly updating the policy with~\eqref{eq:policy_update} is fragile. We instead update $\theta$ by minimizing,
\begin{equation}
\label{eq:policy_update_ours}
\begin{aligned}
&L_{\text{PPO-Lagrangian}}\\ &:= -\mathbb E_{(s,a)\sim \pi_{\theta_\text{old}}}\left[\min\left(rA(s,a), g(\epsilon, A(s,a))\right)\right]\\ &+ \frac{\lambda}{\alpha} \mathbb E_{\xi_\text{T}\sim \pi_{\theta_\text{old}}}\left [\min\left({r} (\eta-\widetilde D(\xi_\text{T}))^+, g(\delta, (\eta - \widetilde D(\xi_\text{T}))^+) \right)\right]
\end{aligned}
\end{equation}

\begin{equation}
    r = \frac{\pi_\theta(s|a;\alpha)}{\pi_{\theta_{\text{old}}}(s|a;\alpha)}
\end{equation}

\begin{equation}
    g(\epsilon, A) =\begin{cases}
    (1+\epsilon)A,\quad \text{if  } A \ge 0\\
    (1-\epsilon)A, \quad \text{if } A < 0,
    \end{cases}
\end{equation}

\begin{equation}
\label{eq:bootstrap}
    \widetilde D(\xi_T) = \sum_{(a_t, s_t)\in \xi_T} \gamma^t R(s_t,a_t) + \gamma^T V_{M,\pi_{\theta_\text{old}}}(s_T), 
\end{equation}
where, $\theta_\text{old}$ is the parameter of the policy used to collect current on-policy batch, $A$ is the generalized advantage estimates described in Eq 16~\cite{schulman2015high}. $\epsilon$ and $\delta$ are the clipping thresholds for the PPO loss and the Lagrangian respectively. Here,~\eqref{eq:policy_update_ours} augments the PPO clipped surrogate with the clipped Lagrangian term; We use the gradient of ~\eqref{eq:policy_update_ours} to update $\theta$ and use the gradient calculated in ~\eqref{eq:eta} and ~\eqref{eq:lambda} to update $\eta$ and $\lambda$ to reflect the true constraint geometry.  The complete procedure is described in Algorithm~\ref{alg:cppo}. 

The design of the algorithm is motivated by three considerations. First, Monte Carlo gradient estimates from~\eqref{eq:policy_update} suffer from high variance, so we adopt PPO with generalized advantage estimation~\cite{schulman2015high}. Second, large unconstrained steps can cause catastrophic performance drops~\cite{schulman2015trust}; PPO-style clipping prevents extreme tail returns from producing unbounded penalty gradients when $\alpha$ is small. Third, in massively parallel settings~\cite{rudin2022learning}, estimating $D(\xi)$ requires rolling out the entire trajectory until termination. This is inefficient in a parallel setting. Thus, we use a bootstrapped return $\widetilde D(\xi_T)$ to estimate returns from truncated length-$T$ trajectories $\xi_T$~\cite{sutton1998reinforcement}.
\begin{algorithm}
  \caption{Offline: CVaR Policy Optimization}
  \label{alg:cppo}
  \begin{algorithmic}[1]
    \Require hyperparameters $\alpha$, $\epsilon$, $\delta$, $\lambda_\text{max}$ learning rates $lr_\eta, lr_\theta, lr_\lambda$, function $f$ for updating $\beta$, rollout length $T$
    \Ensure policy $\pi_\theta$
    \For{$j = 1,2,\ldots, N_{\mathrm{iter}}$}
      \State Generate $N$ trajectories $\{\xi_T^{(i)}\}^N_{i=1}$ with $\pi_{\theta_\text{old}}$.
      \State Compute $A(s_t, a_t)$ and $\widetilde D(\xi^{(i)}_T)$ $\forall i,\; \forall(s_t, a_t)\in \xi_T^{(i)}$
      \State Calculate $\nabla_\theta L_\text{PPO-Lagrangian}$
      \State $\eta \leftarrow \eta - lr_\eta \cdot \nabla_\eta L$
      \State $\theta \leftarrow \theta - lr_\theta \cdot \nabla_{\theta} L_\text{PPO-Lagrangian}$
      \State $\lambda \leftarrow \lambda + lr_\lambda \cdot \nabla_\lambda L$
      \State $\lambda \leftarrow \min(\max(0,\lambda), \lambda_\text{max})$
      \State $\beta \leftarrow f(\{\xi_T^{(i)}\}^N_{i=1})$
      \State $\theta_\text{old} \leftarrow \theta$ 
    \EndFor
  \end{algorithmic}
\end{algorithm}

\subsection{Bandit-based Online Policy Selection}

Choosing the risk level $\alpha$ trades off safety and performance: very small $\alpha$ can be overly conservative and vice versa. We select $\alpha$ by choosing the policy that maximizes episodic (undiscounted) return in the target environment. Because we lack privileged information (e.g., height maps, friction), we assume only access to the realized reward during interaction, computable online from observations and proprioception, so policy selection reduces to a stochastic bandit problem (Sec.~\ref{sec:problem}).

We use the Empirical-Bernstein UCB~\cite{maurer2009empirical, mnih2008empirical}, which is tighter than Hoeffding-based UCB~\cite{bandit2019} and typically converges faster. At episode $e$, for each arm $k$ we compute.
\begin{equation}
\label{eq:ucb}
\mathrm{UCB}_k(e)
:= \hat \mu_k(e)
\sqrt{\frac{2\,\hat \sigma_k(e)\,\ln\!\bigl(3/\delta_e\bigr)}{N_k(e)}}
\frac{3 R\,\ln\!\bigl(3/\delta_e\bigr)}{N_k(e)},
\end{equation}
and select the arm with the largest UCB value. Here, $N_k(e)$ is the number of pulls of arm $k$, $\hat \mu_k(e) = \frac{1}{N_k(e)}\sum_{l=1}^e X_{k,l}$ and $\hat\sigma_k(e) = \frac{1}{N_k(e)-1}\sum_{l=1}^e(X_{k,l} -\hat\mu_k(e))^2$ are the sample mean and variance of the episodic return of arm $k$, $R$ is the reward range, and $\delta_e=\delta/(K e^2)$. The description of the selection algorithm is provided in Algorithm~\ref{alg:alpha-ucb}.

\begin{algorithm}[t]
\caption{Online: Empirical–Bernstein UCB}
\label{alg:alpha-ucb}
\begin{algorithmic}[1]
\Require policies with different $\alpha$-level$ \{\pi_{\theta_k}(\cdot|\cdot,\alpha_k)\}^K_{k=1}$, reward range $R$, episode length $T$
\State Initialize selection counter $N_k \leftarrow 0$
\For{$k=1$ to $K$}
  \State Run $\pi_{\theta_k}$ for $T$ step; save episodic return $x$
  \State Calculate $\hat \mu_k$ and $\hat \sigma_k$ from history of $x$
\EndFor
\For{$e = K+1$ to $E$}
  \State $\delta_e \gets \delta/(K\,e^2)$
  \For{$k=1$ to $K$}
    \State Calculate $\hat \sigma_k$ and $\hat \mu_k$ from history of $x$
    \State $U_k \gets \text{UCB}\big(\hat\mu_k,\,\hat \sigma_k,\,N_k,\,R,\,\delta_e\big)$
  \EndFor
  \State $k^\star \gets \arg\max_k U_k$
  \State $N_{k^\ast} \leftarrow N_{k^\ast} + 1$
  \State Run policy $\pi_{\theta_k}$ for $T$ step; save episodic return $x$
\EndFor
\end{algorithmic}
\end{algorithm}

\section{Simulation Experiments}

In this section, we validate our RL framework in simulation: Section \ref{sec:alg} studies the training behavior when compared to other robust RL approaches based on CVaR optimization. Section \ref{sec:sim_robust} investigates the tradeoff of different $\alpha$ value when deployed to perturbed environments. 

\textbf{Training Setup} We built our policy based on skrl~\cite{serrano2023skrl}. The actor and critic network are implemented as multi-layer perceptrons with hidden layers of dimension $[256, 256, 256]$ and Exponential Linear Unit (ELU) activation. As the return distribution of the policy improves during training, we adaptively change the CVaR constraint level, $\beta$. We set $\beta$ to the exponential moving average of the current $\text{VaR}_\alpha$ using,
\begin{equation}
\label{eq:beta}
    \beta_{t} \leftarrow 0.3 \text{VaR}^\alpha_{\xi\sim\pi}(\widetilde D(\xi_T)) + 0.7\beta_{t-1}.
\end{equation}
Given that, $\text{VaR}_\alpha \ge \text{CVaR}_\alpha$, this schedule keeps the constraint active, progressively shifting probability mass out of the lower tail. We trained the policy in Isaac Lab~\cite{mittal2023orbit} in a flat environment with reward terms defined in Table~\ref{tab:reward} and domain randomization added to distort the observation and randomize joint gain. We describe the domain randomization parameters in Table~\ref{tab:rand_param}. We also set up a curriculum with gradually increasing command magnitude for stable training. 

\begin{table}[t]
  \centering
  \caption{Reward terms used in training.}
  \label{tab:reward}
  \setlength{\tabcolsep}{6pt}
  \renewcommand{\arraystretch}{1.15}
  \begin{tabular}{@{}l c r@{}}
    \toprule
    \textbf{Reward Term} & \textbf{Definition} & \textbf{Weight} \\
    \midrule
    track linear velocity xy & $\text{exp}({-\|\mathbf{v}_{xy}^{\mathrm{cmd}}-\mathbf{v}_{xy}\|^{2}/\sigma})$ & $1.5$ \\
    track angular velocity yaw& $\text{exp}({-(\omega_{\mathrm{yaw}}^{\mathrm{cmd}}-\omega_{\mathrm{yaw}})^{2}/\sigma})$ & $0.75$ \\
    linear velocity z l2   & $v_{z}^{2}$ & $-1.0$ \\
    angular velocity xy l2  & $\|\omega_{\mathrm{roll,pitch}}\|^{2}$ & $-0.05$ \\
    joint acceleration l2        & $\|\ddot{\mathbf{q}}\|^{2}$ & $-2.5\times 10^{-7}$ \\
    torques l2                   & $\|\boldsymbol{\tau}\|^{2}$ & $-2\times 10^{-4}$ \\
    action rate l2          & $\|\mathbf{a}\|^{2}$ & $-0.05$ \\
    feet air time             & $\sum_{f=1}^{4}\!\left(t_{\mathrm{air},f}-0.5\right)$ & $0.35$ \\
    flat orientation l2 & $\|\phi_{\text{roll,pitch}}\|^2$&$-2.0$\\
    \bottomrule
  \end{tabular}
\end{table}

We trained 2048 agents in parallel for 30,000 steps within the environment. This approximately takes 15 minutes wall clock time on a workstation with 24 Core AMD-5900X CPU with a NVIDIA A5000 GPU. 

\begin{table}[t]
  \centering
  \caption{Domain Randomization Parameters}
  \label{tab:rand_param}
  \setlength{\tabcolsep}{10pt}
  \renewcommand{\arraystretch}{1.15}
  \begin{tabular}{@{}ll@{}}
    \toprule
    \textbf{Randomization Parameter} & \textbf{Range} \\
    \midrule
    joint gain      & $[22,27]$
    \\
    joint damping   & $[0.3,0.7]$
    \\
    noise added to base velocity obs              & $[-0.1,0.1]$
    \\
    noise added to ang velocity obs              & $[-0.2,0.2]$
    \\
    noise added to projected gravity & $[-0.05, 0.05]$ \\
    noise added to joint velocity obs & $[-1.5, 1.5]$\\
    \bottomrule
  \end{tabular}
\end{table}

\subsection{Algorithm Performance}
\label{sec:alg}

In these experiments, we demonstrate the training performance of our method, Algorithm~\ref{alg:cppo}, by comparing with multiple risk-aware RL algorithms. We compare our method to \textbf{PPO}~\cite{schulman2017proximal}; \textbf{CPPO}~\cite{ying2022towards}, a simplified variant of ours without clipping or bootstrapped–return estimation; \textbf{return-capped PPO}~\cite{mead2025return}, which directly optimizes Problem~\ref{eq:problem1}; and \textbf{DPPO}, which optimizes the same objective by shaping the value-distribution with QR-DQN~\cite{dabney2018distributional} and SR($\lambda$)~\cite{nam2021gmac}. All baselines use the same training setup, hyperparameters, and architecture; for DPPO the critic head is replaced with a quantile head. We use CVaR at level $\alpha=0.25$.

\begin{figure}[t]
\centering
\includegraphics[width=\columnwidth]{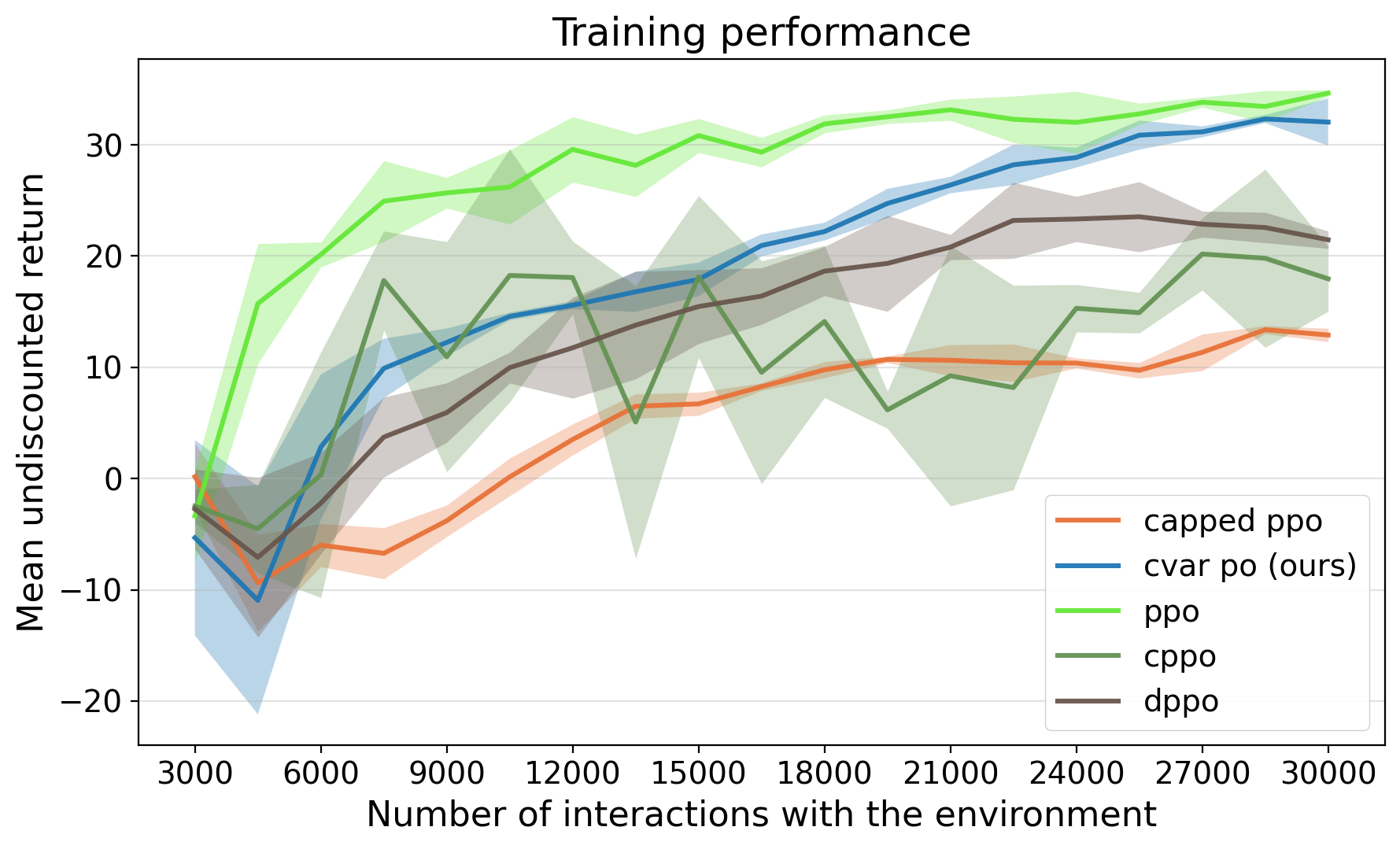}
\caption{Training performance at $\alpha=0.25$. Curves show mean undiscounted return versus timesteps of environment interactions; solid lines are the mean over 10 runs trained for $30{,}000$ timesteps, shaded regions denote $\pm1$ standard deviation. Evaluations are in the training environment.}
\label{fig:training-performance}
\end{figure}

Figure~\ref{fig:training-performance} shows that our method learns more slowly at the start, due to optimizing the Lagrangian term, but attains comparable performance as PPO at the end. CPPO is unstable and slower to converge, motivating the clipping and bootstrapping components in our design. Capped PPO and DPPO, which both directly target the CVaR objective, display similar behavior for much of training; because they de-emphasize high-return trajectories, they tend to learn less from successes.

\begin{figure*}[!t]
  \centering
    \includegraphics[width=\textwidth,keepaspectratio]{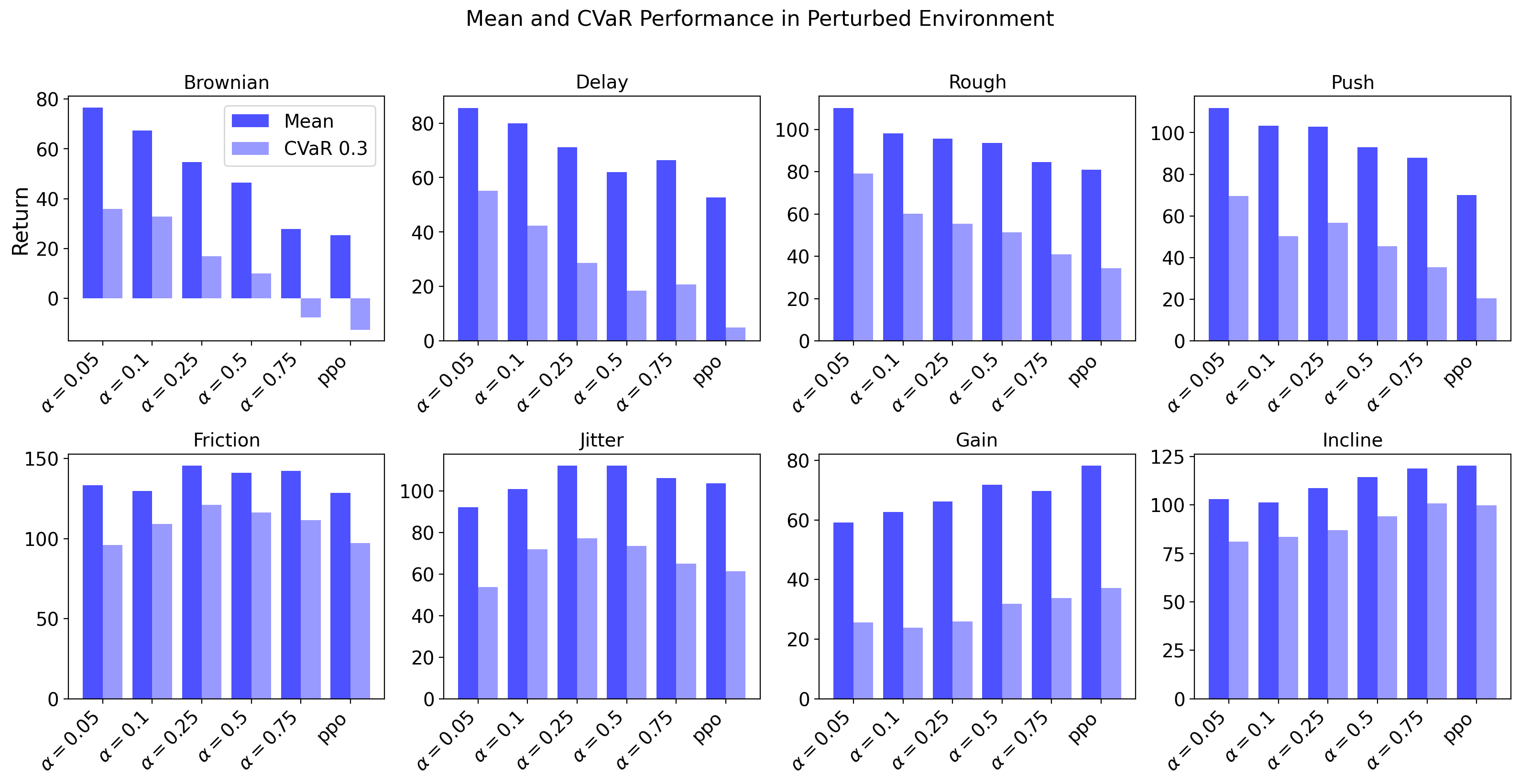}
  \caption{Performance of with different $\alpha$-level. We trained policies different $\alpha$-level and test their performances in perturbed environments. The descriptions of the environments are available at Table~\ref{tab:env}. For each $\alpha$-level we trained 10 policies using different seeds, and we evaluate each policy with 1000 parallel simulations and report the mean return over 6000 timesteps. We reset the environment and randomize the state of robots every 100 timesteps}
  \label{fig:robustness}
\end{figure*}

\subsection{Robustness Under Disturbances}
\label{sec:sim_robust}
In the previous subsection, we compared the performance of our method, Algorithm~\ref{alg:cppo}, to various baselines within the training distribution. Next, we study the robustness–performance trade-off across risk levels to show the need for online policy selection in unknown environments. We train six policies: five CVaR policies with $\alpha\in\{0.05,0.1,0.25,0.5,0.75\}$ and a PPO baseline. Each policy is evaluated in eight out-of-distribution environments that emulate shifts in low-level control, sensing/estimation, terrain, and collisions (Table~\ref{tab:env}).

As seen in Figure~\ref{fig:robustness}, across the first four settings (Brownian, Delay, Rough, Push), smaller $\alpha$ yields higher mean and tail performance; in Brownian and Delay, the $\alpha=0.05$ policy achieves nearly twice the mean and CVaR of PPO, consistent with robustness guarantees for small $\alpha$. The trade-off is evident in the remaining settings: $\alpha=0.25$ performs best on Friction and Jitter, while PPO leads on Gain and Incline, two environment conditions closest to training. Thus, neither the most conservative policy nor nominal PPO is uniformly optimal. This motivates our design of an online and adaptive policy selection scheme, Algorithm~\ref{alg:alpha-ucb}.

\begin{table}[t]
  \centering
  \caption{Environment Description}
  \label{tab:env}
  \setlength{\tabcolsep}{10pt}
  \renewcommand{\arraystretch}{1.15}
  \begin{tabular}{@{}ll@{}}
    \toprule
    \textbf{Environment} & \textbf{Description} \\
    \midrule
    Brownian      & Brownian noise with drift $10^{-3}$ added to linear  \\ &velocity observations
    \\
    Delay   & Joint command delayed by 0.05s
    \\
    Rough              & Rough terrain with maximum height difference 10cm
    \\
    Push              & Random xy linear velocity added to base every 5s
    \\
    Friction & Randomize static friction in range [0.3,0.7]  \\
    Jitter & Replace current observations with past \\ &observations with probability 0.6 \\
    Gain & Randomize gain in range [10, 50] \\
    Incline & The ground plane has a 20 degree incline \\
    \bottomrule
  \end{tabular}
\end{table}

\section{Hardware Experiments}

We now validate our offline training and online policy selection on a Unitree Go2 quadruped. Section \ref{sec:real} compares the real-world performance at different values of $\alpha$. Section \ref{sec:bandit} validates the performance of our policy selection algorithm based on UCB bandits.

\textbf{Hardware Setup} We deployed our policy on a Unitree Go2 Edu robot. The policy sends target joint position commands at 50Hz and the commands are tracked by the robot's PD controller with proportional gain, $k_p = 25$, and derivative gain, $k_d = 0.5$.  We use the OptiTrack motion capture system to estimate the base linear velocities and base angular velocities, and we retrieve the reading from proprioception sensor (the joint position, joint velocity, and feet contact force obtained from the encoder in each robot joint) and gyroscope for the state estimation. We distorted the observations with the same domain randomization as in simulation. 


\subsection{Sim-to-real Robustness}
\label{sec:real}

\begin{figure*}[!t]
  \centering
  \includegraphics[width=\textwidth,keepaspectratio]{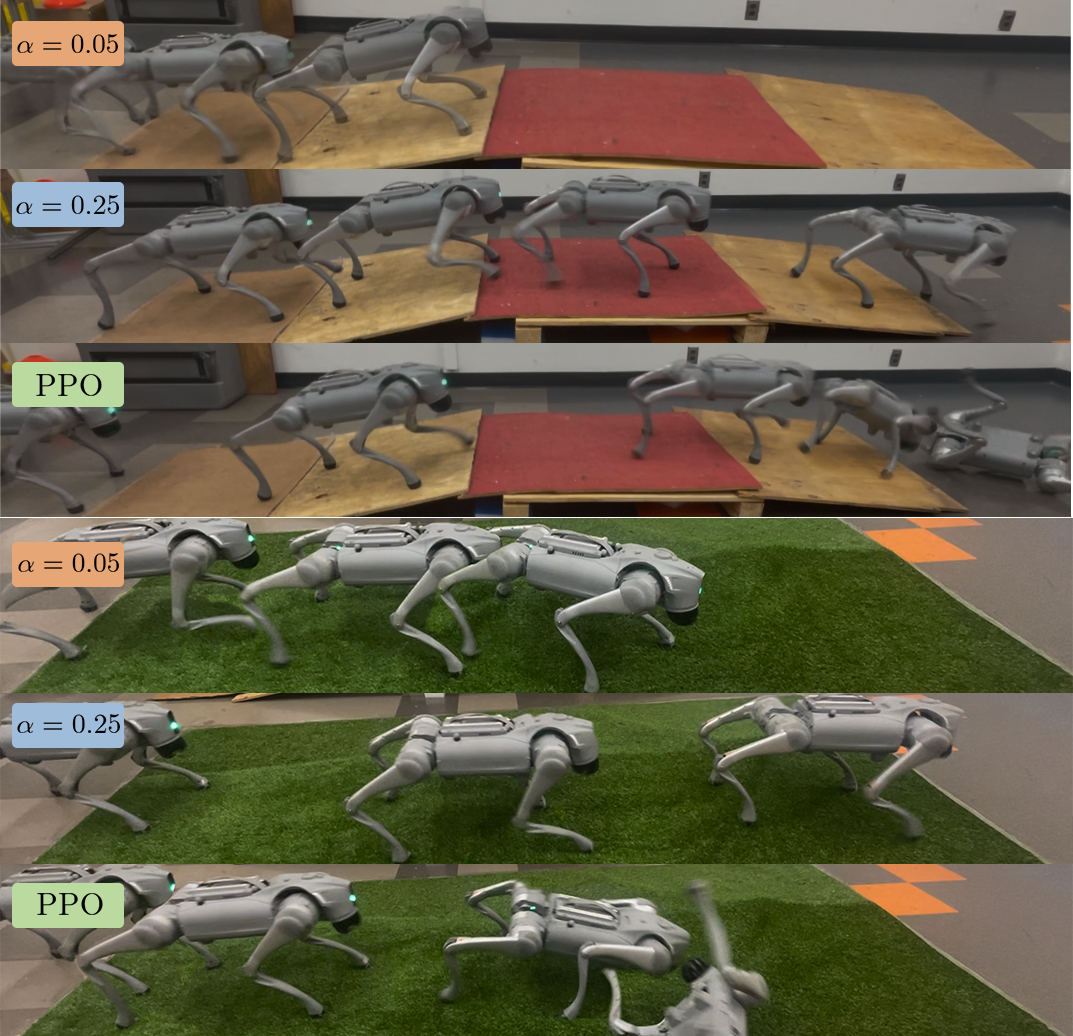}
  \caption{We commanded the robot to walk at full speed ($v=1\,\text{m/s}$) on both ramp and grass terrains. The $\alpha=0.05$ policy did not advance despite forward commands, either when climbing the ramp or when a foot became trapped in a soccer cone hidden beneath the grass. In contrast, the $\alpha=0.25$ policy adapted a stable gait with larger steps, enabling the robot to step over hidden cones. The PPO baseline moved faster but was more prone to losing balance, particularly when descending the ramp or when one of its foot hits the soccer cone.}
  \label{fig:ramp}
\end{figure*}

We deployed the our policies trained at $\alpha\!=\!0.05$ and $\alpha\!=\!0.25$, and the PPO policy on a Unitree Go2 to assess real-world performance. Three settings were used. \textbf{Flat:} the robot was commanded to walk back and forth on flat ground. We did not introduce any disturbances in this environment. \textbf{Ramp:} the robot was commanded to traverse an unseen ramp at constant speed $v=0.7\mathrm{m/s}$, introducing observation shifts, external forces, and unexpected contacts. \textbf{Grass:} the robot walked on turf with $11$ cm soft soccer cones hidden beneath to simulate rough grassland; a trial was successful if the robot remained on its feet while tracking commands for 80s. We measured success rate, time-to-live (TTL), and reward rate. We conducted 15 experiments for each $\alpha$-level in each environment. Results are in Table~\ref{tab:alpha-ramp-grass}; qualitative behaviors at $v=1\,\mathrm{m/s}$ are illustrated in Fig.~\ref{fig:ramp}.

On flat ground, we observe that the PPO moves at a faster and uniform gait compared to $\alpha=0.05$ and $\alpha=0.25$, giving larger return. On the ramp, the $\alpha\!=\!0.05$ policy advanced with shorter steps and was unable to climb 30 \% of the times (Fig.~\ref{fig:ramp}), which explains the longer completion times. The $\alpha\!=\!0.25$ policy developed a stable trotting gait and achieved the best overall performance (highest success and per step reward), with most failures due to occasional drift off the ramp when not precisely following commands. PPO can be unstable and loss balance when unexpected contact happens, resulting in the lowest success rate. On grass, we observed a similar trade-off: PPO achieved the highest reward rate but was fragile to disturbances; $\alpha\!=\!0.05$ attained the best success rate and longest TTL by maintaining stability in this unseen condition.

\begin{table}[t]
  \centering
  \setlength{\tabcolsep}{4pt}
  \renewcommand{\arraystretch}{1.1}
  \caption{Policy performance across three environments.}
  \label{tab:alpha-ramp-grass}
  \begin{tabular}{l l ccc}
    \toprule
    \textbf{Experiment} & \textbf{Policy} & \textbf{Time/TTL (s)} & \textbf{Succ (\%)} & \textbf{Reward/sec} \\
    \midrule
    \multirow{3}{*}{Flat}
      & $\alpha=0.05$ & $18.0 \pm 5.0$ & 100 & $1.05 \pm 0.20$ \\
      & $\alpha=0.25$ & $15.0 \pm 7.0$ & 100 & $1.31 \pm 0.25$ \\
      & PPO           & \bm{$12.0 \pm 6.0$} & 100 & \bm{$1.35 \pm 0.30$} \\
    \midrule
    \multirow{3}{*}{Ramp}
      & $\alpha=0.05$ & $19.0 \pm 8.0$ & 55 & $0.95 \pm 0.35$ \\
      & $\alpha=0.25$ & $12.5 \pm 6.4$ & \bm{$70$} & \bm{$1.32 \pm 0.32$} \\
      & PPO           & \bm{$7.4 \pm 6.3$} & 30 & $1.18 \pm 0.46$ \\
    \midrule
    \multirow{3}{*}{Grass}
      & $\alpha=0.05$ & \bm{$77.0 \pm 9.4$} & \bm{$85$} & $1.01 \pm 0.13$ \\
      & $\alpha=0.25$ & $59.2 \pm 28.1$ & 60 & \bm{$1.14 \pm 0.32$} \\
      & PPO           & $64.2 \pm 36.0$ & 55 & $1.09 \pm 0.44$ \\
    \bottomrule
  \end{tabular}
\end{table}

\subsection{Online Policy Selection and Regret}
\label{sec:bandit}

In this experiment, we evaluate the convergence of the multi-arm bandit when deploying the robot across three testing environments. Using Algorithm~\ref{alg:alpha-ucb} with an episode length of $T=10$, the bandit selects among $\alpha=0.05$, $\alpha=0.25$, and PPO as the robot walks on flat ground, traverses a ramp, and moves over grass. We conducted seven trials per environment and recorded both the collected rewards and the selection history.

\begin{figure}[t]
\centering
\includegraphics[width=\columnwidth]{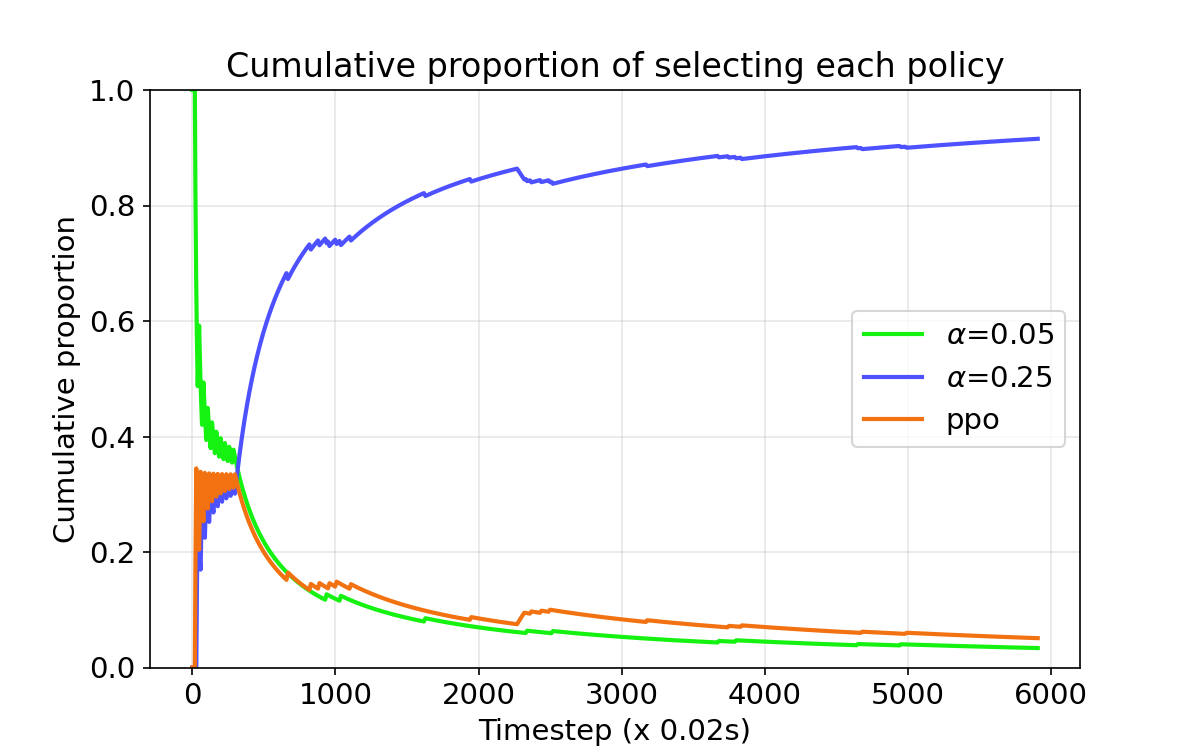}
\caption{Cumulative probability of selecting each policy in one ramp experiment. Selection of policy with $\alpha = 0.25$ quickly dominates after 2000 timesteps, which is roughly 1 minute wall clock time.}
\label{fig:cumulative_selection}
\end{figure}

\begin{figure}[t]
\centering
\includegraphics[width=\columnwidth]{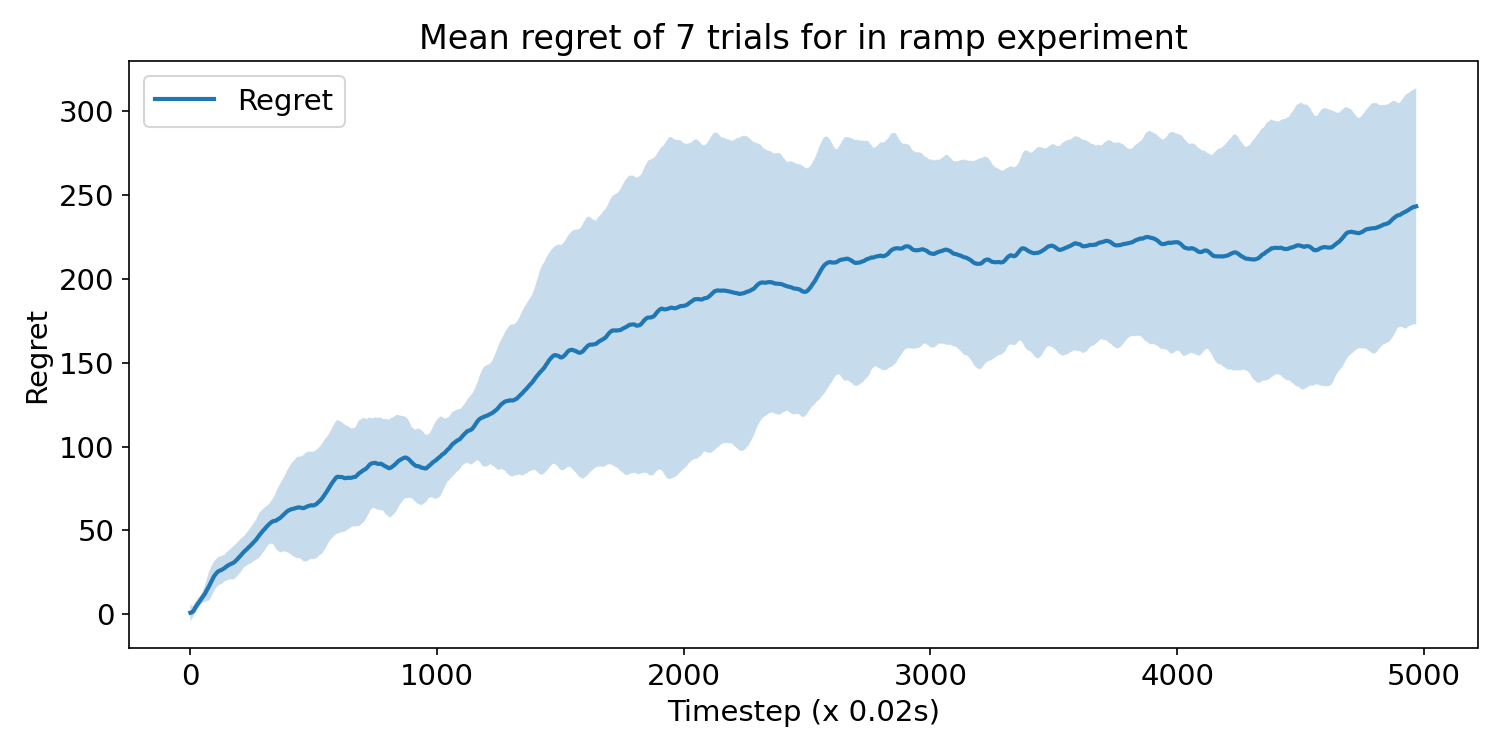}
\caption{Mean regret over time of the 7 ramp experiment. We calculate the regret using~\eqref{eq:problem2} for each experiment, and we plot the mean (solid blue) and $\pm$ 1 standard deviation (light blue region)}
\label{fig:regret}
\end{figure}

Table~\ref{tab:bandit-selection-prob} shows how the selection probabilities evolve over time. In the Flat and Ramp experiments, the probability of choosing the highest-reward policy steadily increases and converges to 1 within 5000 timesteps (about 2 minutes). Figure~\ref{fig:cumulative_selection} illustrates that this convergence often happens quickly after the warm-up rounds in the first 500 timesteps, and Figure~\ref{fig:regret} confirms that the mean regret converges to a constant. By contrast, in the Grass experiment the convergence is less clear because the two higher-reward policies have similar means and high variance. Here, the bandit reliably rules out $\alpha=0.05$ but splits its choices between the remaining two.

\begin{table}[t]
  \centering
  \setlength{\tabcolsep}{4pt}
  \renewcommand{\arraystretch}{1.1}
  \caption{Selection probability by timestep for three policies.}
  \label{tab:bandit-selection-prob}
  \begin{tabular}{l l cccc}
    \toprule
    \textbf{Experiment} & \textbf{Policy} & \textbf{1250} & \textbf{2500} & \textbf{3750} & \textbf{5000} \\
    \midrule
    \multirow{3}{*}{Flat}
      & $\alpha=0.05$ & 0.10 & 0.11 & 0.10 & 0.00 \\
      & $\alpha=0.25$ & 0.20 & 0.19 & 0.10 & 0.00 \\
      & PPO            & 0.70 & 0.70 & 0.80 & 1.00 \\
    \midrule
    \multirow{3}{*}{Ramp}
      & $\alpha=0.05$ & 0.07 & 0.05 & 0.05 & 0.00 \\
      & $\alpha=0.25$ & 0.38 & 0.75 & 0.77 & 1.00 \\
      & PPO            & 0.54 & 0.20 & 0.17 & 0.00 \\
    \midrule
    \multirow{3}{*}{Grass}
      & $\alpha=0.05$ & 0.15 & 0.10 & 0.00 & 0.00 \\
      & $\alpha=0.25$ & 0.50 & 0.53 & 0.75 & 0.53 \\
      & PPO            & 0.34 & 0.37 & 0.25 & 0.47 \\
    \bottomrule
  \end{tabular}
\end{table}

\section{Conclusion and Limitations}

We present a method for generating policies with different risk profiles using CVaR-constrained reinforcement learning. To adapt risk levels online, we introduce a simple UCB-bandit selection mechanism that adjusts based on the observed rewards during task execution. Our risk-aware policies show robustness across environments with diverse disturbances, and the bandit algorithm converges quickly to the best-performing policy.

Two limitations remain: (1) we assume stable average returns, which may fail in nonstationary settings—this could be mitigated with windowed UCB at the cost of extra tuning; (2) the bandit relies on the same reward used in training, overlooking execution risk and thus not favoring safer low-reward policies. Incorporating explicit risk measures into deployment remains an open direction. 

\section{Acknowledgements}

We thank our UCLA colleagues—Allen Emmanuel Binny, Debajyoti Chakrabarti, and Qizhao Chen—for helpful feedback and support. We also thank David Martinez for OptiTrack system assistance and Professor Ankur Mehta for insightful guidance on improving the project.










\begingroup
\raggedright

\endgroup


\begin{thebibliography}{10}
\providecommand{\url}[1]{#1}
\csname url@samestyle\endcsname
\providecommand{\newblock}{\relax}
\providecommand{\bibinfo}[2]{#2}
\providecommand{\BIBentrySTDinterwordspacing}{\spaceskip=0pt\relax}
\providecommand{\BIBentryALTinterwordstretchfactor}{4}
\providecommand{\BIBentryALTinterwordspacing}{\spaceskip=\fontdimen2\font plus
\BIBentryALTinterwordstretchfactor\fontdimen3\font minus \fontdimen4\font\relax}
\providecommand{\BIBforeignlanguage}[2]{{%
\expandafter\ifx\csname l@#1\endcsname\relax
\typeout{** WARNING: IEEEtran.bst: No hyphenation pattern has been}%
\typeout{** loaded for the language `#1'. Using the pattern for}%
\typeout{** the default language instead.}%
\else
\language=\csname l@#1\endcsname
\fi
#2}}
\providecommand{\BIBdecl}{\relax}
\BIBdecl

\bibitem{di2018dynamic}
J.~Di~Carlo, P.~M. Wensing, B.~Katz, G.~Bledt, and S.~Kim, ``Dynamic locomotion in the mit cheetah 3 through convex model-predictive control,'' in \emph{2018 IEEE/RSJ international conference on intelligent robots and systems (IROS)}.\hskip 1em plus 0.5em minus 0.4em\relax IEEE, 2018, pp. 1--9.

\bibitem{mittal2023orbit}
M.~Mittal, C.~Yu, Q.~Yu, J.~Liu, N.~Rudin, D.~Hoeller, J.~L. Yuan, R.~Singh, Y.~Guo, H.~Mazhar, A.~Mandlekar, B.~Babich, G.~State, M.~Hutter, and A.~Garg, ``Orbit: A unified simulation framework for interactive robot learning environments,'' \emph{IEEE Robotics and Automation Letters}, vol.~8, no.~6, pp. 3740--3747, 2023.

\bibitem{rudin2022learning}
N.~Rudin, D.~Hoeller, P.~Reist, and M.~Hutter, ``Learning to walk in minutes using massively parallel deep reinforcement learning,'' in \emph{Conference on robot learning}.\hskip 1em plus 0.5em minus 0.4em\relax PMLR, 2022, pp. 91--100.

\bibitem{tobin2017domain}
J.~Tobin, R.~Fong, A.~Ray, J.~Schneider, W.~Zaremba, and P.~Abbeel, ``Domain randomization for transferring deep neural networks from simulation to the real world,'' in \emph{2017 IEEE/RSJ international conference on intelligent robots and systems (IROS)}.\hskip 1em plus 0.5em minus 0.4em\relax IEEE, 2017, pp. 23--30.

\bibitem{shi2024robust}
J.~Shi, C.~Bai, H.~He, L.~Han, D.~Wang, B.~Zhao, M.~Zhao, X.~Li, and X.~Li, ``Robust quadrupedal locomotion via risk-averse policy learning,'' in \emph{2024 IEEE International Conference on Robotics and Automation (ICRA)}.\hskip 1em plus 0.5em minus 0.4em\relax IEEE, 2024, pp. 11\,459--11\,466.

\bibitem{schulman2015high}
J.~Schulman, P.~Moritz, S.~Levine, M.~Jordan, and P.~Abbeel, ``High-dimensional continuous control using generalized advantage estimation,'' \emph{arXiv preprint arXiv:1506.02438}, 2015.

\bibitem{schulman2017proximal}
J.~Schulman, F.~Wolski, P.~Dhariwal, A.~Radford, and O.~Klimov, ``Proximal policy optimization algorithms,'' \emph{arXiv preprint arXiv:1707.06347}, 2017.

\bibitem{ying2022towards}
C.~Ying, X.~Zhou, H.~Su, D.~Yan, N.~Chen, and J.~Zhu, ``Towards safe reinforcement learning via constraining conditional value-at-risk,'' \emph{arXiv preprint arXiv:2206.04436}, 2022.

\bibitem{schneider2024learning}
L.~Schneider, J.~Frey, T.~Miki, and M.~Hutter, ``Learning risk-aware quadrupedal locomotion using distributional reinforcement learning,'' in \emph{2024 IEEE International Conference on Robotics and Automation (ICRA)}.\hskip 1em plus 0.5em minus 0.4em\relax IEEE, 2024, pp. 11\,451--11\,458.

\bibitem{wang2000class}
S.~S. Wang, ``A class of distortion operators for pricing financial and insurance risks,'' \emph{Journal of risk and insurance}, pp. 15--36, 2000.

\bibitem{bellemare2017distributional}
M.~G. Bellemare, W.~Dabney, and R.~Munos, ``A distributional perspective on reinforcement learning,'' in \emph{International conference on machine learning}.\hskip 1em plus 0.5em minus 0.4em\relax PMLR, 2017, pp. 449--458.

\bibitem{dabney2018distributional}
W.~Dabney, M.~Rowland, M.~Bellemare, and R.~Munos, ``Distributional reinforcement learning with quantile regression,'' in \emph{Proceedings of the AAAI conference on artificial intelligence}, vol.~32, no.~1, 2018.

\bibitem{dabney2018implicit}
W.~Dabney, G.~Ostrovski, D.~Silver, and R.~Munos, ``Implicit quantile networks for distributional reinforcement learning,'' in \emph{International conference on machine learning}.\hskip 1em plus 0.5em minus 0.4em\relax PMLR, 2018, pp. 1096--1105.

\bibitem{mead2025return}
H.~Mead, C.~Costen, B.~Lacerda, and N.~Hawes, ``Return capping: Sample-efficient cvar policy gradient optimisation,'' \emph{arXiv preprint arXiv:2504.20887}, 2025.

\bibitem{schulman2015trust}
J.~Schulman, S.~Levine, P.~Abbeel, M.~Jordan, and P.~Moritz, ``Trust region policy optimization,'' in \emph{International conference on machine learning}.\hskip 1em plus 0.5em minus 0.4em\relax PMLR, 2015, pp. 1889--1897.

\bibitem{achiam2017constrained}
J.~Achiam, D.~Held, A.~Tamar, and P.~Abbeel, ``Constrained policy optimization,'' in \emph{International conference on machine learning}.\hskip 1em plus 0.5em minus 0.4em\relax PMLR, 2017, pp. 22--31.

\bibitem{miryoosefi2019reinforcement}
S.~Miryoosefi, K.~Brantley, H.~Daume~III, M.~Dudik, and R.~E. Schapire, ``Reinforcement learning with convex constraints,'' \emph{Advances in neural information processing systems}, vol.~32, 2019.

\bibitem{yang2023neural}
R.~Yang, G.~Yang, and X.~Wang, ``Neural volumetric memory for visual locomotion control,'' in \emph{Proceedings of the IEEE/CVF conference on computer vision and pattern recognition}, 2023, pp. 1430--1440.

\bibitem{kumar2021rma}
A.~Kumar, Z.~Fu, D.~Pathak, and J.~Malik, ``Rma: Rapid motor adaptation for legged robots,'' \emph{arXiv preprint arXiv:2107.04034}, 2021.

\bibitem{yang2021learning}
R.~Yang, M.~Zhang, N.~Hansen, H.~Xu, and X.~Wang, ``Learning vision-guided quadrupedal locomotion end-to-end with cross-modal transformers,'' \emph{arXiv preprint arXiv:2107.03996}, 2021.

\bibitem{tamkin2019distributionally}
A.~Tamkin, R.~Keramati, C.~Dann, and E.~Brunskill, ``Distributionally-aware exploration for cvar bandits,'' in \emph{NeurIPS 2019 Workshop on Safety and Robustness on Decision Making}, 2019.

\bibitem{mnih2008empirical}
V.~Mnih, C.~Szepesv{\'a}ri, and J.-Y. Audibert, ``Empirical bernstein stopping,'' in \emph{Proceedings of the 25th international conference on Machine learning}, 2008, pp. 672--679.

\bibitem{contextual}
D.~Bouneffouf, I.~Rish, and C.~Aggarwal, ``Survey on applications of multi-armed and contextual bandits,'' in \emph{2020 IEEE Congress on Evolutionary Computation (CEC)}, 2020, pp. 1--8.

\bibitem{bandit2019}
A.~Slivkins, \emph{Introduction to Multi-Armed Bandits}.\hskip 1em plus 0.5em minus 0.4em\relax Foundations and Trends in Machine Learning, 2019.

\bibitem{rockafellar2000optimization}
R.~T. Rockafellar, S.~Uryasev \emph{et~al.}, ``Optimization of conditional value-at-risk,'' \emph{Journal of risk}, vol.~2, pp. 21--42, 2000.

\bibitem{ma2025dsac}
X.~Ma, J.~Chen, L.~Xia, J.~Yang, Q.~Zhao, and Z.~Zhou, ``Dsac: Distributional soft actor-critic for risk-sensitive reinforcement learning,'' \emph{Journal of Artificial Intelligence Research}, vol.~83, 2025.

\bibitem{bertsekas1997nonlinear}
D.~P. Bertsekas, ``Nonlinear programming,'' \emph{Journal of the Operational Research Society}, vol.~48, no.~3, pp. 334--334, 1997.

\bibitem{JMLR:v18:15-636}
\BIBentryALTinterwordspacing
Y.~Chow, M.~Ghavamzadeh, L.~Janson, and M.~Pavone, ``Risk-constrained reinforcement learning with percentile risk criteria,'' \emph{Journal of Machine Learning Research}, vol.~18, no. 167, pp. 1--51, 2018. [Online]. Available: \url{http://jmlr.org/papers/v18/15-636.html}
\BIBentrySTDinterwordspacing

\bibitem{sutton1998reinforcement}
R.~S. Sutton, A.~G. Barto \emph{et~al.}, \emph{Reinforcement learning: An introduction}.\hskip 1em plus 0.5em minus 0.4em\relax MIT press Cambridge, 1998, vol.~1, no.~1.

\bibitem{maurer2009empirical}
A.~Maurer and M.~Pontil, ``Empirical bernstein bounds and sample variance penalization,'' \emph{arXiv preprint arXiv:0907.3740}, 2009.

\bibitem{serrano2023skrl}
\BIBentryALTinterwordspacing
A.~Serrano-Muñoz, D.~Chrysostomou, S.~Bøgh, and N.~Arana-Arexolaleiba, ``skrl: Modular and flexible library for reinforcement learning,'' \emph{Journal of Machine Learning Research}, vol.~24, no. 254, pp. 1--9, 2023. [Online]. Available: \url{http://jmlr.org/papers/v24/23-0112.html}
\BIBentrySTDinterwordspacing

\bibitem{nam2021gmac}
D.~W. Nam, Y.~Kim, and C.~Y. Park, ``Gmac: A distributional perspective on actor-critic framework,'' in \emph{International Conference on Machine Learning}.\hskip 1em plus 0.5em minus 0.4em\relax PMLR, 2021, pp. 7927--7936.

\end{thebibliography}
\end{document}